\documentclass{article}
\usepackage{PRIMEarxiv}
\usepackage{amsthm}
\usepackage{amsmath}
\usepackage{algorithm}
\newtheorem{theorem}{Theorem}
\usepackage{booktabs}
\usepackage{algpseudocode}
\algblock{Input}{EndInput}
\algnotext{EndInput}
\algblock{Output}{EndOutput}
\algnotext{EndOutput}
\newcommand{\Desc}[2]{\State \makebox[2em][l]{#1}#2}

\usepackage[utf8]{inputenc}
\usepackage[greek, english]{babel}

\usepackage{tikz}
\newtheorem{remark}{Remark}
\usepackage[utf8]{inputenc} 
\usepackage[T1]{fontenc}    
\usepackage{url}            
\usepackage{booktabs}       
\usepackage{amsfonts}       
\usepackage{nicefrac}       
\usepackage{microtype}      
\usepackage{lipsum}
\usepackage{fancyhdr}       
\usepackage{graphicx}       
\graphicspath{{media/}}     

\usepackage[pdftex, colorlinks]{hyperref}

\title{Resolving the Exploitation-Exploration Dilemma in Evolutionary Algorithms: A Novel Human-Centered Framework}

\author{
  Ehsan Shams \\
  Department of Mathematics and Computer Science\\
  Faculty of Science, Alexandria University \\
  Alexandria, 21526, Egypt\\
  \texttt{ehsan.ali@alexu.edu.eg}\\ Mar, 2025 
}

\begin{document}
\newfloat{Procedure}{tbp}{lop}
\maketitle

\begin{abstract}
Evolutionary Algorithms (EAs) are widely employed tools for complex search and optimization tasks; however, the absence of an overarching operational framework that permits a systematic regulation of the exploration-exploitation tradeoff—critical for efficient convergence—restricts the full actualization of their potential, leading to the so-called exploration-exploitation dilemma in algorithm design. A systematic resolution to this dilemma requires: (1) independent yet coordinated control over exploration and exploitation, and (2) an explicit, computationally feasible, adaptive regulation mechanism. The current,  almost decentralized, traditional parameter tuning-centeric approach—lacks the foundation to satisfy these requirements under encoding-imposed structural constraints.

We propose a Human-Centered Two-Phase Search (HCTPS) framework, in which the actualization of (1) and (2) is enabled through an external configuration variable—the \emph{Search Space Control Parameter} (SSCP). As the sole control knob of HCTPS, the SSCP centralizes exploration adjustments, sparing users from micromanaging traditional parameters with unintelligible interdependencies. To this construct, the human user serves as a meta-parameter, adaptively steering the regulatory process via SSCP adjustments. 

We prove that the HCTPS strictly surpasses the current approach in terms of search space coverage without disrupting the EAs’ inherent convergence mechanisms, demonstrate a concrete instantiation of it—using the Genetic Algorithm as the underlying heuristic on a suite of global benchmark unconstrained optimization problems, provide a through assessment of the proposed framework, and envision future research directions. Any search algorithm prone to this dilemma can be applied in light of the proposed framework, being algorithm-agnostic by design.
\end{abstract}


\section{Introduction}
\label{sec: intro}
Evolutionary Algorithms (EAs) are a family of stochastic computational models that approximate the processes of genetic variation (variation), selection pressure (natural selection), adaptation to selection pressure (adaptation), and reproductive heritability observed in biological evolution.

Variation refers to the differences in characteristics among organisms within a population. These differences naturally arise due to small changes in inherited traits, which can result from genetic mutations, recombination, or other natural processes. Natural selection refers to the survival challenges imposed by the environment, while adaptation is the process by which genetic variations that enhance an organism’s fitness in its respective environment become more prevalent in a population over generations through natural selection, and reproductive heritability ensures that beneficial adaptations are passed down to subsequent generations. Together, these processes contribute to steering biological systems toward progressively more robust and well-adapted states. 

EAs enable systems of artificial entities—e.g., strings representing candidate solutions to a problem or programs executing specific tasks—to build themselves up over many iterations in response to a selection pressure exerted by a given fitness criterion. First, a random population of artificial entities is generated, and their performance is evaluated according to this fitness criterion which then rewards the fitter entities with a higher survival advantage, allowing them to reproduce through mutation and recombination. Over successive iterations, this evolution ideally yields entities with progressively higher fitness—e.g., solutions that more effectively solve a given problem or programs capable of executing the intended tasks with near-optimal precision—thus filling the available niche in the artificial ecosystem.

Over the years, numerous variants of EAs have been proposed. The primary distinctions between them lie in their representation schemes—how evolving entities are encoded—and the manner in which evolutionary processes are discretized and implemented. Notable examples include Evolution Strategies (ES)(Rechenberg, 1984), Evolutionary Programming (EP), Genetic Algorithms (GAs) (Holland, 1975), and Genetic Programming (GP) (Koza, 1992).

By design, EAs are particularly well-suited for optimization problems, particularly those  characterized by complex, highly nonlinear, and multimodal function landscapes, and where traditional optimization algorithms often struggle. This realization can be traced back to the pioneering work of computer scientists in the 1950s and 1960s, who first explored the potential of evolutionary processes as optimization tools for engineering problems (Mitchell, 1998). While EAs are mostly applied to optimization problems, their applicability extends to a wide range of problem domains across science, engineering, artificial intelligence, arts, music, knowledge discovery, and more. Beasley (2000) overviews some of the application areas of EAs.

The course of adaptation in natural evolutionary systems is a result of maintaining the proper tradeoff between two implicit forces: exploration and exploitation. Exploration is induced by the diversification mechanisms that give rise to genetic variability within a population, while exploitation is induced by natural selection, which systematically capitalizes on genetic variations that enhance organisms’ fitness. An improper tradeoff between these two forces can destabilize systems, and if persistently maintained, may ultimately lead to a total collapse.

Unlike machines, nature is remarkably effective at self-regulating this exploration-exploitation tradeoff, evidenced by the continuation of life on Earth. Therefore, it is often very difficult to engineer artificial evolutionary systems (e.g. EAs) capable of regulating this tradeoff autonomously. This difficulty leads to the so-called exploration-exploitation dilemma. This dilemma, first introduced in the EAs literature by authors of (Eiben \& Schippers, 1998), refers to the challenge of optimally allocating computational resources, in the proper ratio, to the exploration and exploitation mechanisms within the algorithm to ensure efficient evolution with respect to the problem landscape under consideration.

\subsection{Requirements for a Systematic Resolution}
A systematic resolution to this dilemma necessitates a principled framework in which the following are met:

\begin{enumerate}
\item \textbf{Decoupled and independent control}: Exploration and exploitation must be regulated through distinct and independently adjustable mechanisms while maintaining controlled coordination to prevent search instability. This decoupling ensures that adjustments to exploration do not implicitly degrade exploitation or vice versa, thereby preventing entangled tradeoffs that hinder search efficiency.

\item \textbf{Explicit and adaptive regulation}: The framework must provide an explicitly defined, dynamically adjustable exploration control mechanism that can be adapted in response to measurable search progress indicators (e.g., search space coverage, fitness diversity,final algorithm output).

\item \textbf{Theoretical and computational feasibility}: The framework must be algorithm-agnostic, and its internal mechanisms must remain computationally scalable without requiring an exponential increase in representation set size, prohibitively large population sizes, or exhaustive parameter tuning. 
\end{enumerate}

To the best of our knowledge, the current approach practiced in the literature for addressing this dilemma in EAs primarily centers around distributing tuning efforts among traditional parameters, such as the selection of evolutionary operators, operator probabilities, population size, and others. In this context, one often engages in the development of new operators internal to the algorithm and experiment with other various parameter configurations in pursuit of optimizing algorithm search performance. And indeed, while specific tuning configurations can lead to improvements—such as certain EA operators outperforming others in specific contexts— this approach does not amount to a systematic resolution to the dilemma.

The unintelligible interdependencies among these parameters obscure their individual contributions to exploration and exploitation. This prevents independent, controlled modulation of the exploration-exploitation tradeoff, and does not provide a principled basis for adaptive control, thus prohibiting the actualization of requirements (1), and (2). 

Moreover, this approach under encoding-induced structural constraints further restrict the algorithm's ability to enhance exploration without excessive computational overhead, thus prohibiting the actualization of requirement (3). Hence, the need for a new framework that provides the foundations for systematic resolution of the dilemma in question.

\subsection{Contribution}
The primary contributions of this work are as follows:

\begin{enumerate}

\item \textbf{Rigorous background for context:}
We provide background sections on the canonical GA and the canonical EA framework to offer a fresh, rigorous, and stand-alone overview of standard EA mechanics and parameters, which culminate in key insights that motivate and inform the proposed HCTPS framework.
  
\item \textbf{Technical analysis of current approach limitations}: We analyze how the unintelligible interdependencies among traditional EA parameters, coupled with encoding-based structural constraints, prevent principled, adaptive regulation of exploration and exploitation in details.

\item \textbf{Novel Human-Centered Two-Phase Search (HCTPS) framework:}
We introduce a new framework that \emph{externalizes} exploration from the EA’s internal mechanics, placing it under a single, user-driven configuration variable called the \emph{Search Space Control Parameter} (SSCP). We show how this decoupling provides a systematic basis for resolving the exploration–exploitation dilemma.

\item\textbf{Proof of strictly enhanced exploration}: Through a rigorous argument, we prove that HCTPS strictly increases the search space coverage compared to the current approach for an arbitrary EA, ensuring that the exploration-exploitation tradeoff can be modulated without degrading convergence mechanisms.

\item \textbf{Empirical validation}: We demonstrate a concrete instantiation of the HCTPS framework—using canonical GA as the underlying heuristic into a two-phase, human-centered search process on a set of global unconstrained blackbox optimization problems and compare results against the current approach, and we comment on the output.

\item \textbf{Assessment of the proposed framework}:
We evaluate HCTPS in terms of its ability to meet the earlier requirements, conduct a cost-benefit analysis, and discuss wider applicability.
\end{enumerate}

\subsection{Structure of the Paper}
The remainder of this paper is structured to be as self-contained as possible, with an emphasis on clarity, and the use of analogies to facilitate intuitive and comprehensive explanations and is organized around the contributions identified above as follows:

Sections \ref{sec:GA}--\ref{sec:EA} deliver \emph{contribution 1}, while section \ref{sec:CA} \emph{contribution 2}, and section \ref{sec: HCTPS}  \emph{contributions 3 and 4}. Section \ref{sec:experiments} delivers \emph{contribution 5}, and section \ref{sec:assess} \emph{contribution 6}. Finally section \ref{sec:conclusions} summarizes the paper, offers key insights, concludes the work, and presents future research directions.

\section{The Canonical Genetic Algorithm (GA)}
\label{sec:GA}
The first variant of GAs was introduced by Holland in the 1960s and later refined in the 1970s through collaborations at the University of Michigan. It is commonly referred to in the literature as the simple or canonical GA. Like all EAs, GAs are inspired by the Darwinian theory of evolution (Darwin, 1859) but their terminology is derived from natural genetics.

The progression of the GA is organized into six key steps, each implemented through specific stochastic operators that collectively shape its evolutionary cycle. These steps are detailed below.

\subsection{Terminology, Key Steps, and Operators}

\subsubsection*{Step 0. Encoding}
The algorithm begins by generating a set of potential solutions, $S_r$, to the problem at hand of cardinality $N$, chosen uniformly at random from within the feasible solution set $S$. Each solution in $S_r$ is then encoded (or represented) \footnote{Throughout this paper, we use the terms encoding and representation interchangeably} into a finite string of bits of fixed length $L$ \footnote{The length of a string is the number of its constituent bits.}. 

Formally, the encoding process is described by the bijective map:

\begin{center}
ENCODE: $S_r \to X$,
\end{center}

Where $X$ is the set of all possible strings with length $L$ in the chosen representation system. For example, with binary encoding of length $L$, $X$ is the set of all binary strings of length $L$, and the cardinality of $X$ is $|X|=2^L$. The selection of the encoding system—whether binary, ternary, decimal, alphabetic, or another format— depends on the characteristics of the problem and must be selected carefully to ensure semantic accuracy and efficiency in representing the solutions.

\begin{remark}
Alternatively, one may begin with a randomly generated subset of strings of length $L$ from a chosen representation system (e.g., binary, ternary, etc.) and then apply a DECODE function, which should also be bijective, to map these strings back into the original feasible set $S$.
\end{remark}

Terminology:
\begin{itemize}
\item Each encoded string is referred to as \underline{chromosome}.
\item Each bit constituting the chromosome is called a \underline{gene} and a
\underline{population} is a collection of chromosomes.
\item We call the set $X$ the representation set, while the base elements of the representation system (e.g. binary \{0,1\}, ternary \{0,1,2\}, ..etc) are called representative elements.
\end{itemize}

\subsubsection*{Step 1. Evaluation}

Once the population has been encoded, the fitness of each chromosome is evaluated using a non-negative function derived from the main objective function to be optimized called the "fitness function". The fitness function assigns a numerical fitness value to each chromosome, serving as a quantitative measure of its performance and quality within the current population. The idea behind assigning fitness values is to discriminate good from bad quality solutions.

\subsubsection*{Step 2. Selection}

Based on the fitness values, a probability
measure is derived to determine the likelihood of each potential solution (chromosome) being selected as a parent
for reproduction in the next step. The aim here is to reward fitter chromosomes with a survival advantage, that is, a higher probability of contributing their genetic material to the next generation, and to penalize the less fit, making them more likely to die out/get excluded in the process.  This aligns with the ”survival of the fittest” principle.

In nature, however, unfit genes are not always immediately eliminated. Some may become recessive, persisting within the population for extended periods. This persistence allows for the possibility that future environmental changes or mutations may transform these genes into beneficial traits. As such, there is a trade-off between promoting fitter chromosomes with advantageous traits and preserving genetic diversity within the population.

GA designers acknowledge this trade-off by employing selection operators that balance the need for promoting fitness with maintaining population diversity. These operators are typically classified based on how they assign probabilities of selection—commonly referred to as selection pressure—to chromosomes within the population. Common selection operators include: 
\begin{enumerate}
\item  Roulette wheel selection (fitness-proportional selection): Assigns selection probabilities proportional to fitness values (Goldenberg, n.d.). While favoring fitter chromosomes, it can lead to premature convergence if fitness differences are extreme.
\item Ordinal selection (rank-based selection): Assigns selection probabilities based on rank rather than raw fitness values pressure (Baker, 1985). Instead of using fitness directly, individuals are ranked, and probabilities are distributed according to their position. This method ensures a more controlled selection pressure, preventing any single individual from overwhelming the population, thus maintaining genetic diversity.
\end{enumerate}

\subsubsection*{Step 3. Recombination}
In the recombination step, the GA combines bits and pieces of two or more parental chromosomes through the crossover operator to generate new hopefully better offspring chromosomes with a pre-defined probability $p_c$ \footnote{Further details can be found in the next section}. This operator aims to create offspring that inherit desirable traits from their parents while introducing novel combinations of genetic material.

The specific definition and implementation of the crossover operator can vary, and its effectiveness greatly depends on designing a suitable recombination mechanism. Many variants of the crossover operator exist, such as single-point, multi-point, uniform crossover (detailed description of these can be found in (De Jong, 2006); geometric crossover (Michalewicz, 2013), heuristic crossover (Wright, 1991), and partially mapped crossover (PMX) (Goldberg \& Lingle, 1985).

The fundamental principle behind recombination is to ensure that the offspring are distinct from the parents. This allows the algorithm to explore new regions of the solution space while preserving key features of the parent chromosomes that contribute to fitness.

\subsubsection*{Step 4. Mutation}

Unlike the crossover operator which operates on two or more parental chromosomes, the mutation operator \underline{locally but randomly}, modifies a chromosome according to a pre-defined probability $p_m$ which is often small.

There are many variations of the mutation operators such as bit-flipping mutation (for binary representation), uniform mutation, non-uniform mutation, Gaussian mutation and Cauchy mutation; a comprehensive overview of these operators can be found in (Liu \& Han, 2003). The common aspect shared by all the variations is the involvement of one or more changes that are made to the genes of the chromosomes. In other words, mutation performs a random walk in the vicinity of a candidate solution, adding a level of stochasticity and diversity to the GA.

\subsubsection*{Step 5. Replacement}
The final step in the GA is replacement, where the algorithm determines how to incorporate newly generated chromosomes into the next generation. This process is carried out using a replacement operator, which plays a crucial role in shaping the diversity and convergence rate of the algorithm. There are several variants of the replacement operator, each influencing the evolutionary process differently. Among the most common are: elitist replacement, generation-wise replacement, and steady-state replacement; a comprehensive overview can be found in (Eiben \& Smith, 2015).

\subsection{Parameters}
\label{subsec:par}
A parameter in the context of an algorithm is a configuration variable that influences the algorithm's behaviour, performance, and convergence properties, whose value remains fixed during the algorithm's execution. 

\textbf{Formal definition}:
Let \(\mathcal{A}\) be an algorithm operating on a input space \(\mathcal{X}\) and producing outputs in the space \(\mathcal{Y}\), represented as:

\[
\mathcal{A}: \mathcal{X} \times \Theta \rightarrow \mathcal{Y}.
\]

Where \(\theta\) is a parameter to \(\mathcal{A}\) and \(\Theta\) denotes the space of all possible parameter values, then \(\theta\)'s role is to define the conditions under which the mapping applies.

Parameters can be classified based on their temporal behavior, updating strategy, and impact on algorithm performance.

\begin{itemize}
\item \textbf{Temporal behaviour}: Parameters’ values in an algorithm can either remain static throughout the execution of all the iterations or change dynamically based on the algorithm’s progress according to some rule. 

\begin{itemize}
\item Dynamic parameter: Some parameters are adjusted between iterates in response to feedback from the algorithm. For example, the mutation probability $p_m$ in a GA may be modified if the algorithm is not converging effectively—increasing to enhance exploration or decreasing to focus on exploitation.
\item Static parameter: Other parameters are defined before execution and remain unchanged throughout all the iterates. For instance, the population size in a GA is typically set before the algorithm starts running and stays constant throughout the whole run.
\end{itemize}
\item \textbf{Updating strategy}: Parameters during the algorithm's run may either self-update based on feedback from the algorithm or be set and updated by the human user (human-centered). In the first, we say such parameter is mechanically-centered, and in the latter human-centered.  Examples:
\begin{itemize}
\item \textbf{Mechanically-centered parameter}: The mutation probability $p_m$ in a GA can be dynamically adjusted based on the fitness improvement of the best solution found. If the fitness improves rapidly, $p_m$ may be decreased to focus efforts on exploitation; otherwise, it may be increased to encourage exploration. 
\item \textbf{Human-centered parameter}: These parameters remain fixed during execution. Examples include the maximum number of iterations and the selection of genetic operators (e.g., crossover and mutation type).
\item \textbf{Impact on algorithm behavior}: Parameters vary in their impact on the algorithm’s performance. Some parameters have a stronger influence, meaning even small changes can significantly alter the search behavior, while others require larger adjustments to produce noticeable effects.
\end{itemize}

\begin{remark}
The parameter definition presented here applies not only to computational algorithms, which are a specific class of mappings or operators, but more generally to any operator or a system of operators. In the latter, parameters define the structural and operational constraints under which the system functions.
\end{remark}

Below, we analyze the traditional GA parameters in regards to their impact on the exploration-exploitation of the search space.

\begin{enumerate}

\item \textbf{Population size}
Population size is a critical parameter for balancing the exploration and exploitation dynamics of a GA. A small population size limits exploration by reducing diversity, which increases the risk of premature convergence to sub-optimal solutions (Roeva et al., 2013). Conversely, a larger population size promotes exploration by sampling a broader range of the solution space, reducing the likelihood of stagnation in local optima. However, this comes at the cost of slower convergence, which may hinder the algorithm's exploitation of promising solutions.

\item \textbf{Crossover probability $p_c$}. The crossover operator is the primary search mechanism GAs and one of their defining features compared to other EAs (Tian, 2001). The crossover probability $p_c$ controls the likelihood that two parent solutions will undergo crossover to produce offspring. A higher $p_c$ accelerates convergence by intensifying the recombination of genetic material, potentially leading to faster exploitation of promising solutions. However, excessive crossover may reduce population diversity, hindering the algorithm's ability to explore the solution space effectively. 

\textbf{Example}: If $p_c=1$ and the population consists of $N$ chromosomes, every pair of chromosomes will undergo crossover. In contrast, if  $p_c=0.5$, there is a $50 \%$ chance for any given pair to participate in crossover. 

\item \textbf{Mutation probability $p_m$}. The mutation probability $p_m$ dictates the likelihood of a chromosome undergoing random gene-modifications. Higher $p_m$ enhance exploration by introducing randomness, which helps the algorithm escape local optima and discover new regions of the solution space. However, excessive mutation can disrupt favorable solutions, destabilizing the algorithm and reducing convergence efficiency. Conversely, lower $p_m$ promotes exploitation by preserving high-quality solutions but may decrease diversity, increasing the risk of premature convergence.

\textbf{Example}: If $p_m=0.1$, there is a $10 \% $ chance for each chromosome in the population to undergo mutation. In a population of $N$ chromosomes, approximately $\frac{N}{10}$ of chromosomes are expected to be selected for mutation. However, due to the stochastic nature of the process, the actual number of mutations may vary.

\item \textbf{Selection of genetic operators}. The choice of GA operator—including selection, crossover, mutation, and replacement—determines the descriptive content (or the logic) of the mechanism by which individuals are chosen for reproduction, how genetic material is recombined, and how it is altered to produce new solutions. 

\textbf{Example}: Consider optimizing a function where the objective is to minimize a complex, multimodal surface. Descriptively different selection operators will have different impact on the exploitation-exploration dynamics.

\begin{itemize}
\item Elitism selection: In each generation, the best-performing solutions (e.g., top 10\%) are guaranteed to survive into the next generation. This promotes exploitation by preserving high-quality solutions but risks reducing diversity if overused, potentially causing premature convergence to local optima.

\item Tournament selection: Groups of solutions compete, and the best solution in each group is selected for reproduction. This balances exploration and exploitation by favoring fitter solutions while still allowing less fit ones a chance to propagate, maintaining diversity.

\item Random selection: Individuals are chosen without regard to fitness. This maximizes exploration, ensuring broad sampling of the solution space, but sacrifices exploitation, making it unlikely to focus on improving promising solutions.
\end{itemize}

\item \textbf{Representation}.
The representation of solutions determines how the genetic information is encoded and decoded for the problem at hand. The choice of representation directly affects the search space, the feasibility of generated solutions, and the algorithm's exploration and exploitation dynamics. A well-suited representation ensures efficient exploration and accurate mapping of candidate solutions.

\textbf{Example}: The Traveling Salesman Problem (TSP) requires finding the shortest route that visits a set of cities and returns to the starting city. Two common representations are:

\begin{itemize}
\item Binary Representation: Here each gene represents whether a city is visited or not. For example, a chromosome might be represented as [1, 0, 1, 1, 0] indicating that the first, third, and fourth cities are visited. The order of the cities is not explicitly represented in this representation.

\item Permutation Representation: Here, we choose a permutation representation where each gene represents the order in which the cities are visited. For example, a chromosome might be represented as [3, 1, 4, 2, 5] indicating that the cities are visited in the order of 3, 1, 4, 2, and 5.

\end{itemize}

Impact on the exploration-exploitation dynamics:

\begin{itemize}
\item The choice of binary representation limits the search space to a subset of all possible permutations which negatively affects the algorithm's exploration. Since the order of the cities is not explicitly represented, the GA needs to incorporate additional mechanisms to generate valid tours. This representation may require more complex decoding mechanisms and genetic operators to ensure the generated solutions are feasible tours. However, it can be advantageous in reducing the search space complexity and improving computational efficiency, especially for large TSP instances.

\item The choice of permutation representation allows for a direct mapping of the chromosome to a valid tour. The search space includes all possible permutations of the cities, which is a larger search space compared to the binary representation. However, the permutation representation simplifies the decoding process and ensures that all generated solutions are valid tours. This representation may require more computational resources due to the larger search space, but it allows for a more straightforward exploration of different tour configurations.
\end{itemize}

\item \textbf{Number of fitness function evaluations}. Increasing the number of evaluations allows the algorithm to gather more information about the problem space, enhancing its ability to explore diverse regions and identify potential solutions in a meaningful way. At the same time, this provides opportunities to refine high-performing solutions, improving accuracy and solution quality through exploitation. However, excessive evaluations can significantly increase computational costs, slowing down the algorithm, while too few evaluations may result in premature convergence and sub-optimal solutions.

An analogy can be drawn to a traveler lost in a dark forest with a limited supply of oil for their lamp. Adding more oil brightens the lamp, allowing for better and meaningful navigation and exploration of the forest. However, the faster the oil is used, the less is left for future needs. Similarly, increasing fitness evaluations enhances both exploration and exploitation quality but requires careful balancing to optimize the use of computational resources. Striking this balance is essential to ensure the algorithm is efficient while achieving high-quality results.

\end{enumerate}
\end{itemize}

\section{The Canonical Evolutionary Algorithm (EA)}
\label{sec:EA}
So far we discussed in detail the conceptual framework and key steps of the canonical GA. However, this framework is not exclusive to GAs; it applies to all EAs, meaning that the same six-step structure in (\ref{sec:GA}) applies to all EAs. In the introduction of this paper, we briefly mentioned several notable examples of EAs, and outlined that they differ in two key areas: Representation schemes and the approximation of evolutionary processes employed.

In the next subsection, we give a summary to how these differences manifest among various EAs, followed by key insights into EA design and broad implications.

\subsection{Key Differences among EAs}
EAs can be distinguished based on two fundamental design aspects:

\begin{enumerate}
\item Representation schemes: 
\begin{itemize}
\item Genetic algorithms (GA): Use bit-string encoding or real-valued representations.
\item Evolution strategies (ES): Employ real-valued vectors, making them more suited for continuous optimization.
\item Genetic programming (GP): Uses tree-based structures, enabling the evolution of symbolic expressions or programs.
\end{itemize}

\item Approximation of evolutionary processes:
\begin{itemize}
\item Selection mechanisms: GAs often rely on fitness-proportionate or tournament selection, whereas ES employs deterministic selection.
\item Reproduction mechanisms: GAs primarily use crossover and bitwise mutations, ES applies Gaussian perturbations, and GP modifies tree structures via subtree recombination.
\item Replacement mechanisms: GAs typically replace the entire population per generation, while ES and GP often incorporate elitist or steady-state replacement strategies to maintain diversity.
\end{itemize}
\end{enumerate}

\subsection{Some Key Insights}
\label{subsec: KI}
\begin{enumerate}
\item Efficient search navigation and convergence in an EA fundamentally relies on achieving a properly balanced exploration-exploitation dynamic relative to the problem environment.
\item The traditional internal operators (e.g., selection, crossover, mutation, replacement) and parameters (e.g., population size, crossover probability, mutation rate) collectively define the EA's inherent potential for exploration and exploitation. 
\item The exploration and exploitation processes driven by such traditional operators and parameters, are inherently conflicting. Optimizing one process often diminishes the other.
\item The structure of an EA tells us that the algorithm operates and moves within the representation set $X$. The cardinality of $X$ is $|X|=K^L$ where $K$ is the number of representative elements and $L$ is the length of the string. 
\item The structure of EAs is abstract enough to give the user flexibility in defining their own operators and parameters, enabling customization and adaptation to specific problem domains.
\item It remains an active area of research to explore and identify new parameters that influences the optimization performance of algorithms in noticeable magnitudes with reasonable tuning requirements.
\end{enumerate} 

\section{Managing the Exploration-Exploitation Dilemma in EAs: The Current Approach}
\label{sec:CA}

As highlighted in the introduction section, the current approach predominantly focus on adjusting traditional EA parameters (e.g., evolutionary operators and population size). While these tuning strategies can offer incremental gains under certain conditions, they do not amount to a systematic resolution of the exploration-exploitation dilemma—largely due to the unintelligible interdependencies among these parameters and the resulting inability to \underline{decouple exploration from exploitation}.


Here, we provide a more detailed analysis and  offer a more concrete exploration of why and how the current approach's limitation persist. To illustrate them clearly, let us consider the case of maximizing exploration in an \textit{idealized setting} in which the algorithm never revisits any point in the search space during a single execution:

As discussed in sections (\ref{sec:GA} \& \ref{sec:EA}): 

\begin{enumerate}
\item EAs operate within the representation set $X$, which defines the space of possible encoded solutions, and where the whole movement of the algorithm occurs. 
\item The set $X$ is in one-to-one correspondence with a proper subset of the whole search cube $I$ under the DECODE function. Formally, if $f$ is the DECODE function, then $X$ is in one-to-one correspondence with $f(X)$ where $f: X \to f(X) \subset I$.
\end{enumerate}

Now, enhancing the exploration potential of an EA requires satisfying two necessary conditions: (1) increasing the size of the representation set $X$ and (2) increasing the population size $N$. As stated in subsection (\ref{subsec: KI}), the size of the representation set is $K^L$, where $K$ is the base of the representation system (e.g., binary (2), ternary (3), decimal (9), etc.) and $L$ is the length of the encoded string representing each  potential solution. So, increasing the size of the representation set$X$, for some fixed representation system, implies increasing the value of $L$.

Assume that the algorithm evaluates $N$ solutions per iteration, and given the assumption that no point in the search will be revisited (which is in practice, not guaranteed to happen) this would imply the need for $m$ number of iterations where $m$ satisfies $\left| m - \frac{K^L}{N} \right| \leq \epsilon$ and $\epsilon = \min \left\{ y \in \mathbb{Z^+} \mid \left|m - \frac{K^L}{N}\right| \leq y \right\}$ for the algorithm to explore the whole representation set.

For instance, in binary representation-based GAs, the size of the representation set is $2^L$. If $L$ is set to be 20 then we have $2^{20}= 1048576$ different binary strings available to be explored. If $N=100$ then, ideally we would need a prohibitively large number of iterations, $m=10486$ iterations, (or slightly smaller) per run- under the assumption that no point will be visited twice -to explore the whole representation set (or a reasonable portion of it) and with that, we may never reach the best solutions.

It can be inferred that with the addressed approach, enhancing the exploration potential not only entails satisfying conditions (1) and (2) but also demands meticulous tuning of internal EA operators (e.g. crossover + mutation) to facilitate the exploration of new search points as much as possible and minimize the risk of getting trapped in a narrow and sub-optimal region of the solution space but without risking it behaving like a random algorithm.

In this, several challenges arise immediately: 

\begin{itemize}
\item The first challenge lies in the significant computational burden incurred when trying to meet conditions (1) and (2). Increasing the size of the representation set and the population requires substantial computational resources, which can limit the efficiency of the algorithm.

\item The second challenge stems from the inherent unintelligibility of the interdependencies between internal EA parameters and operators which further complicates the understanding and prediction of their combined effects on the exploitation-exploration trade-off thereby, making their tuning process difficult.

\item The third challenge is that even if conditions (1) and (2) are met, and a selection of good EA operators was employed, there are no guarantees that the algorithm will reach promising solutions. It is possible that some promising niches may not be included within the bounds of $f(X)$, leading to the risk of overlooking optimal solutions.
\end{itemize}

\textbf{Take away message}: By considering idealized scenario in which no point in the search space is revisited, one sees why simply scaling up population size or adjusting operator parameters still fails to yield a principled mechanism for exploration—especially as it omits any notion of how to preserve exploitation performance.

Given these limitations, we adopt a less-traveled path: rather than viewing exploration and exploitation as irreconcilable forces or favoring one at the expense of the other, we attempt to reconcile them under a new, unifying framework.

\section{A Novel Human-Centered Two-phase Search Framework (HCTPS)}
\label{sec: HCTPS}

\subsection{Framework Definition and Human-Centered Role}
Rather than increasing the size of $X$ and $N$ or direct all efforts to fine-tuning traditional EA parameters (such as mutation and crossover in GAs) to enhance search exploration, we propose a human-centered search framework. This framework dynamically relocates the representation set’s co-domain under $f$ within the search space (i.e., it adaptively shifts the mapping $f$ to sufficiently distinct subsets of $I$). This ensures an increased exploration of new regions within $I$ without compromising the algorithm's ability to exploit promising solutions while also eliminating the inefficiencies of the traditional approach.

In subsection (2.2), we formally defined a parameter as a fixed quantity that influences the behavior of a system without being an explicit variable in execution. We further categorized parameters based on how they are tuned, distinguishing between:

\begin{itemize}
\item Human-centered parameters, which are manually adjusted by the user.
\item Mechanically-centered parameters, which are autonomously updated by the algorithm's internal mechanics.
\end{itemize}

To the proposed search framework, the human user acts as a \textit{meta \footnote{The prefix "meta-" originates from the Greek word \textgreek{ μετά} (meta), meaning "beyond," "after," or "about." In modern usage, it generally signifies a higher level of abstraction or self-reference}-parameter}, meaning they do not directly engage in the execution of the search but instead govern the selection, adaptation, and regulation of its key parameters—most notably, the Search Space Control Parameter (SSCP)—which defines the structure and evolution of the search process.

By the end of this section, the role of the human user as a meta-parameter will become more concrete.

\subsection{Two-Phase Structure of HCTPS}
This framework is structured as a two-phase search process, consisting of:
\begin{enumerate}

\item \textbf{Global Search Phase}: In this phase, a broad exploration of the entire search cube $I$ is conducted. In this, the algorithm's inherent exploration-exploitation potential is distributed across the whole search cube. However, this phase alone may overlook certain promising regions or niches within $I$, necessitating a subsequent localized search to address this limitation. 

\item \textbf{Local Search Phase}: In this phase, a sequence of selected sub-cubes $\{ I_i \}$ from the original search cube $I$ (where $i=1,2,\dots,n$) undergoes further intensive sequential exploration. In this, the algorithm focuses its inherent exploration-exploitation potential over each sub-cube of the generated sequence.

The rigorous implementation of the local search phase is achieved through the introduction and utilization of the search space size control parameter (SSCP) which serves as a defining mechanism for the selection and exploration of the $n$ specific sub-cubes within $I$. 
\end{enumerate}

\subsubsection{The Search Space Size Control Parameter (SSCP)}
The SSCP is an external configuration variable to an EA that represents a selection of a sequence of sub-cubes $\{I\}_{i=1}^n \subset I$ for targeted exploration during the local search phase.

The SSCP follows a human-centered tuning strategy. The following steps outline how the user specifies, updates, and refines the SSCP:

\begin{enumerate}

\item \textbf{Rule-specification and sequence generation}: Before the local search phase begins, the user defines a rule to generate the sequence of sub-cubes $\{I\}_{i=1}^n$. Subsequently, the user generates the $n$ terms of this sequence according to the specified rule, where each term corresponds to a localized region of interest that the algorithm will intensively explore.
\item \textbf{Adaptation and expansion}: During the local search phase, it is up to the user to expand the sequence by introducing new sub-cubes , or modify the sequence-generative rule dynamically based in response to real-time insights.
\end{enumerate}

\subsubsection{HCTPS Procedure}
The following structured procedure outlines the general framework of the two-phase search process with SSCP integration. 

The specific implementation details vary based on the algorithm \( g \) used and the characteristics of the output data, including whether it constitutes a single solution set, a multi-solution set (such as a Pareto set in case of multi-objective optimization), or another form.
\newpage
\begin{Procedure}

\caption{HCTPS Procedure (Structured Steps)}
\label{algo:HCTPS}
\begin{algorithmic}
\Input
  \Desc{$I$: Global search cube}{}
  \Desc{$g$: An arbitrary EA}{}
  \Desc{$k$: Number of iterations per execution of $g$}{}
  \Desc{$T_{\text{satisfactory}}$: Stopping criterion based on user satisfaction}{}
\EndInput
\vspace{0.25cm}
\Output
  \Desc{$S_{\text{best}}$: The best solution(s) found during the entire search process up to $T_{\text{satisfactory}}$}{}
\EndOutput

\vspace{0.25cm}
\State \textbf{Step 1: Global Search Phase:}
\State \quad a) Run $g$ across the entire search cube $I$ to perform a broad exploration experiment for $k$ number of iterations.
\State \quad b) Identify and store the best solution(s) found in this phase as $S_{\text{global}}$.

\State \textbf{Step 2: Local Search Phase}
\State \quad a) Define a sequence-generative rule for selecting sub-cubes \( \{I_i\}_{i=1}^{n} \subset I \)
\State \quad b) Generate these \( n \) sub-cubes in the sequence according to this rule
\State \quad c) For each selected sub-cube \( I_i \):
\State \quad \quad i) Execute \( g \) over \( I_i \) for \( k \) iterations to conduct a local search.
\State \quad \quad ii) Identify the best solution(s) within \( I_i \), store as \( S_i \)

\State \textbf{Step 3: Additional Sub-Cube Selection}
\While{$T_{\text{satisfactory}}$ is not met}
\State \quad a) Generate $n_{\text{additional}}$ new sub-cubes according to either the predefined rule or an updated rule
\State \quad b) Execute step 2c for each new sub-cube
\EndWhile

\State \textbf{Step 4: Output Best Solution(s)}
\State \quad a) Aggregate results from all exploration searchs in Steps 1 and 2:
\State \quad \quad $S_{\text{HCTPS}} = S_{\text{global}} \cup \bigcup_{i=1}^{n+n_{\text{additional}}} S_i$
\State \quad b) Identify the best solution(s) $S_{\text{best}}$ from $S_{\text{HCTPS}}$
\State \quad c) Return $S_{\text{best}}$
\end{algorithmic}
\end{Procedure}

The value of $n_{\text{additional}}$ here represents the number of additional terms in the sequence that can be generated after the $n$ exploration experiments are carried out in real-time in step 2. This allows the decision-maker to dynamically adjust the exploration based on insights gained, problem complexities, and satisfaction level.

\begin{remark}
In the HCTPS framework, the user focuses all of their tuning efforts on one the SSCP.
\end{remark}

\newpage

\subsection{Theoretical Guarantee: HCTPS Ensures Increased Exploration}
\label{subsec: hctps}
\begin{theorem}
\label{thm: hctps}
Let \( I \subset \mathbb{R}^d \) be the search cube, and let \( X \) be the representation set such that an EA maps \( X \) to an \textbf{effective search region} \( I' \subset I \) via a fixed decoding function \( f: X \to I' \). 

Define two search approaches:
\begin{itemize}
    \item \textbf{Traditional:} The EA is applied using a fixed mapping \( f \), meaning the search is restricted to a fixed region \( I' \subset I \).
    \item \textbf{HCTPS:} The search space is divided into sub codomains \( \{I_i\}_{i=1}^{n} \), and \( f \) is dynamically relocated to different sub codomains over successive runs.
\end{itemize}

Then, \textbf{HCTPS guarantees greater exploration coverage than the traditional one}, meaning the probability of covering more of the search space under HCTPS is strictly greater than under the traditional approach.
\end{theorem}

\begin{proof}
Let $ x^* $ be an optimal solution (or a high-quality region containing $ x^*$) in $ I $.

In the traditional approach, the probability of reaching $ x^* $ is given by:
\[
P_{\text{trad}} = P(x^* \in I') = \int_{I'} p(x) dx,
\]
where $ p(x) $ is the probability density function of the search distribution within $I$, which describes the likelihood of sampling different points.

Assume that $p(x)$ is approximately uniform within $I'$, which is reasonable under general EA sampling assumptions. Then

\[
P_{\text{trad}} \approx \frac{|I'|}{|I|} P_{\max},
\]
where:
\begin{itemize}
\item $ P_{\max}$ is the probability of discovering $ x^*$ if the full space $I$ were searched exhaustively with a uniform sampling strategy.
\item $|I|$, $|I'|$ represents the volumes (or measures) of $I$ and $I'$ respectively.
\item $\frac{|I'|}{|I|}$ represents the fraction of the search space that is covered by the EA.
\end{itemize}

Since $ |I'| \ll |I|$, it follows that:
\[
P_{\text{trad}} \ll P_{\max}.
\]
This means that in the traditional approach, the probability of discovering an optimal region or a solution is fundamentally limited by the fixed and often small search region $I'$.

In the local phase of the HCTPS, instead of being confined to a single fixed region $ I'$, the user selects a sequence of $ n$ subregions (subcubes) $ \{I_i\}_{i=1}^{n} \subset I$.
Each execution of the EA occurs within a different $I_i$. This modifies the probability of finding $x^*$ as follows:

\[
P_{\text{local}} = 1 - \prod_{i=1}^{n} (1 - P_{I_i}),
\]
where $P_{I_i}$ is the probability of $x^*$ being located in subcube $I_i$.

Since each $ I_i$ is sufficiently distinct (i.e., covers new unexplored regions), we approximate $P_{I_i}$:
\[
P_{I_i} \approx \frac{|I_i|}{|I|} P_{\max}.
\]
And substitute in the earlier equation:
\[
P_{\text{local}} \approx 1 - \prod_{i=1}^{n} \left(1 - \frac{|I_i|}{|I|} P_{\max} \right).
\]

This reflects the increasing probability of finding $x^*$ as more subcubes are explored.

\textbf{Comparing $P_{\text{trad}}$ and $ P_{\text{local}}$}

Since $ P_{\text{trad}} \approx \frac{|I'|}{|I|} P_{\max}$, we compare it with $ P_{\text{local}}$.

For \textbf{small} $\frac{|I_i|}{|I|} P_{\max} $, we use the first-order approximation:
\[
\prod_{i=1}^{n} \left(1 - \frac{|I_i|}{|I|} P_{\max} \right) \approx 1 - \sum_{i=1}^{n} \frac{|I_i|}{|I|} P_{\max}.
\]
Thus, we obtain:
\[
P_{\text{local}} \approx \sum_{i=1}^{n} \frac{|I_i|}{|I|} P_{\max}.
\]

Since the selected sequence of subcubes covers a total region of volume: $\sum_{i=1}^{n} |I_i| > |I'| $, we conclude that:
\[
P_{\text{local}} > P_{\text{trad}}.
\]
And because $P_{\text{HCTPS}}= P_{\text{trad}} + P_{\text{local}}$, then $P_{\text{HCTPS}}> P_{\text{trad}}$.\\

\textbf{Conclusion}. Since $ P_{\text{HCTPS}} > P_{\text{trad}} $, it follows that HCTPS guarantees superior search coverage and increased exploration compared to the traditional approach. This result holds as long as the selected $ \{I_i\}$ provides sufficient diversity.

Thus, \textbf{HCTPS is mathematically guaranteed to enhance exploration while maintaining inherent exploitation driven by the algorithm's internal mechanics} $\square$
\end{proof}

\subsection{Visualizing the HCTPS with a Helicopter Analogy}
Imagine you are a photographer and you want to capture a breathtaking landscape with a camera equipped with a fixed zoom capability. You take a helicopter to get a bird's-eye view of the entire scenery from a height of 3000 feet. You admire the vastness of the scenery but realize that crucial details are missing from this altitude.
Thereby, you decide to zoom in to be able to explore specific portions of the landscape. Lowering the helicopter, you hover closer to the ground, focusing your camera on different regions to capture the intricacies of each area. As you lower the helicopter and hover closer to the ground while changing your position across the x-y plane, adjusting your camera’s focus on different regions, you will be able to capture new details that were previously hidden from your previous vantage point.

By moving your helicopter \textit{strategically} to various positions in the x-y-z plane, your camera will be able to capture diverse perspectives of the landscape. Each new position offers a unique view, revealing different aspects of the scenery that were previously hidden from other vantage points, allowing you to compose a more faithful representation of the scenery, compared to the one and only 3000 feet view photoshoot. 

This dynamic approach intelligently compensates for the camera's fixed zoom capability through the human-centered adjustment of the helicopter's position, allowing exploration of the landscape from various positions and angles.
In this scenario, the camera's fixed zoom capability is analogous to the algorithm’s inherent exploitation-exploration fixed potential. The 3000 feet view is analogous to the initial search phase (the global phase) in our proposed approach and this is where traditional EAs stop at thereby, only able to explore the search space from a fixed perspective.

The human-centered strategic position-changing of the helicopter is analogous to the role of the decision-maker in directing the local search phase in \ref{algo:HCTPS} where the active participation of the photographer in the strategic adjustment of the helicopter's position in the x-y-z plane is akin to the active participation of decision-maker in adjusting and tuning the SSCP by generating the sequence of sub-cubes to be explored in the local search phase according to some rule.

\section{GA-integrated HCTPS: Experimental Demonstration}
\label{sec:experiments}

The objective of this experimental study is threefold:
\begin{enumerate}
\item Illustrate the implementation of the HCTPS: Show how an EA can be adapted to the two-phase HCTPS framework in practice
\item Demonstrate SSCP tuning: Provide an example into how the user configures and refines the SSCP for a particular problem instance
\item Dominant-parameter illustration: Illustrate how the SSCP is the most dominant parameter, in effect, to effective exploration regulation. 
\end{enumerate} 

Toward goals (1) and (2), we instantiate the procedure (\ref{algo:HCTPS}) using a \emph{canonical GA} as our base algorithm $g$. We evaluate this GA-integrated HCTPS (HCTPS-GA) on 14 black-box, unconstrained optimization problems collected from (Chen et al., 2014; Yang, 2010; Dieterich \& Hartke (2012); Tan, 2016), focusing exclusively on their 30-dimensional versions.

For (3), we compare the proposed framework against the current one (i.e, we compare the performance of the GA integrated into the HCTPS framework (HCTPS-GA) against the standalone GA). See remark (3) in section (\ref{sec: HCTPS}).

These problems exhibit varying levels of difficulty and topological complexity. For instance, functions like F1 (Bent Cigar) are relatively simple and serve as a baseline for performance comparison, while others, such as F10 (Rastrigin) and F11 (Griewank), present significant challenges due to their multimodal nature and the presence of numerous local optima. Moreover, the set contains functions that are specifically designed to be computationally expensive such as F14 (Ackley), thereby simulating real-world scenarios where evaluations of the objective function are costly (Crico et al., 2020), and providing a realistic framework for assessing the effectiveness of various exploration and exploitation EA strategies to navigate complex landscapes (Vaidya et al., 2020). Detailed discussions of these functions' properties are provided in (Plevris \& Solorzano, 2022; Chen et al., 2014).

Our GA operates with a maximum limit of exact objective function evaluations per run, defined as \( \textit{feval} = 50 \times n \), where \( n \) is the dimensionality of the problem. For the 30-dimensional case, \( \textit{feval} = 1500 \). The search cube for all problems is \(I=[-100, 100]^{30}\), which includes the global minimum for each test problem as well. The GA employs binary encoding, the canonical two-point crossover, and binary tournament selection. Statistical results are derived from 20 independent runs.

All experiments in this research were conducted using MATLAB 2022a (The MathWorks, Inc.) on a personal computer running Microsoft Windows 11, equipped with an 11th Gen Intel(R) Core(TM) i7-1165G7 processor @ 2.80 GHz and 16 GB of RAM. Table (\ref{tab:HCTPS-GA-RES}) presents the results of the GA integrated into the HCTPS framework (HCTPS-GA), while Tables (\ref{tab:F1-F2}, \ref{tab:F3-F4}, \ref{tab:F6-F7}, \ref{tab:F7-F8}) provides a side-by-side comparison between the standalone GA and HCTPS-GA.

\begin{table}[htbp]
\resizebox{\textwidth}{!}{
\begin{tabular}{@{}llllllll@{}}
\toprule
\textbf{ID} &
  \textbf{Function Name} &
  \textbf{Mean} &
  \textbf{Best} &
  \textbf{Worst} &
  \textbf{Median} &
  \textbf{St. Dev} &
  \textbf{Execution Time of 20 Runs (s)} \\ \midrule
F1  & Bent Cigar                & 4.27603E-38 & 2.37593E-38 & 5.79992E-38 & 4.23246E-38 & 8.7447E-39  & 1.252549 \\
F2  & Discus                    & 1.16259E-16 & 4.2676E-20  & 5.50934E-16 & 5.60015E-17 & 1.67102E-16 & 1.132544 \\
F3  & Weierstrass               & 6.3873E-05  & 5.45428E-05 & 7.25862E-05 & 6.29509E-05 & 5.08126E-06 & 8.136591 \\
F4  & Modified Schwefel         & 0.008579635 & 0.006739378 & 0.011169685 & 0.008389775 & 0.001468878 & 1.256719 \\
F5  & Katsuura                  & 1.13712E-10 & 9.5848E-11  & 1.26901E-10 & 1.14559E-10 & 7.79472E-12 & 5.830386 \\
F6  & HappyCat                  & 2.840347319 & 2.840347319 & 2.840347319 & 2.840347319 & 4.07358E-12 & 1.16361  \\
F7  & HGBat                     & 0.500000001 & 0.500000001 & 0.500000001 & 0.500000001 & 7.64014E-11 & 1.213726 \\
F8 &
  Expanded Griewank plus Rosenbrock &
  0.459947694 &
  0.459947694 &
  0.459947694 &
  0.459947694 &
  2.64002E-11 &
  1.429998 \\
F9  & Expanded Scaffer's F6     & 4.43632E-11 & 6.56031E-12 & 9.47957E-11 & 3.7284E-11  & 3.07707E-11 & 1.326823 \\
F10 & Rosenbrock's              & 2           & 2           & 2           & 2           & 3.27683E-11 & 1.200114 \\
F11 & Griewank's                & 0           & 0           & 0           & 0           & 0           & 2.515366 \\
F12 & Rastrigin's               & 0           & 0           & 0           & 0           & 0           & 1.292805 \\
F13 & High Conditioned Elliptic & 5.42E-20    & 3.99E-20    & 7.49E-20    & 5.54E-20    & 9.02E-21    & 1.20E+00 \\
F14 & Ackley                    & 1.68E-10    & 1.47E-10    & 1.84E-10    & 1.68E-10    & 1.18E-11    & 1.343571 \\ \bottomrule
\end{tabular}
}
\caption{Statistical Results of HCTPS-GA}
\label{tab:HCTPS-GA-RES}
\end{table}

\begin{table}[htbp]
\centering
\footnotesize

\begin{minipage}[t]{0.48\textwidth}
\centering
\begin{tabular}{@{}llll@{}}
\toprule
\textbf{ID} & \textbf{} & \textbf{HCTPS-GA} & \textbf{GA} \\ \midrule
F1 & \begin{tabular}[c]{@{}l@{}}Mean\\ Best\\ Worst\\ Median\\ St. Dev\end{tabular} & 
\begin{tabular}[c]{@{}l@{}}4.27603E-38\\ 2.37593E-38\\ 5.79992E-38\\ 4.23246E-38\\ 8.7447E-39\end{tabular} & 
\begin{tabular}[c]{@{}l@{}}7.0816E+10\\ 5.7076E+10\\ 9.1041E+10\\ 7.0687E+10\\ 9.5167E+9\end{tabular} \\
F2 & \begin{tabular}[c]{@{}l@{}}Mean\\ Best\\ Worst\\ Median\\ St. Dev\end{tabular} & 
\begin{tabular}[c]{@{}l@{}}1.16259E-16\\ 4.2676E-20\\ 5.50934E-16\\ 5.60015E-17\\ 1.67102E-16\end{tabular} & 
\begin{tabular}[c]{@{}l@{}}1.3119E+8\\ 1.0460E+7\\ 8.4294E+8\\ 1.0484E+7\\ 2.2834E+8\end{tabular} \\
\bottomrule
\end{tabular}
\caption{Performance Comparison (F1-F2)}
\label{tab:F1-F2}
\end{minipage}
\hfill
\begin{minipage}[t]{0.48\textwidth}
\centering
\begin{tabular}{@{}llll@{}}
\toprule
\textbf{ID} & \textbf{} & \textbf{HCTPS-GA} & \textbf{GA} \\ \midrule
F3 & \begin{tabular}[c]{@{}l@{}}Mean\\ Best\\ Worst\\ Median\\ St. Dev\end{tabular} & 
\begin{tabular}[c]{@{}l@{}}6.3873E-05\\ 5.45428E-05\\ 7.25862E-05\\ 6.29509E-05\\ 5.08126E-06\end{tabular} & 
\begin{tabular}[c]{@{}l@{}}6.3440E-5\\ 5.6442E-5\\ 7.0481E-5\\ 6.2658E-5\\ 3.9426E-6\end{tabular} \\
F4 & \begin{tabular}[c]{@{}l@{}}Mean\\ Best\\ Worst\\ Median\\ St. Dev\end{tabular} & 
\begin{tabular}[c]{@{}l@{}}0.008579635\\ 0.006739378\\ 0.011169685\\ 0.008389775\\ 0.001468878\end{tabular} & 
\begin{tabular}[c]{@{}l@{}}5.9077E+3\\ 4.7541E+3\\ 7.5088E+3\\ 5.8122E+3\\ 6.6643E+2\end{tabular} \\
\bottomrule
\end{tabular}
\caption{Performance Comparison (F3-F4)}
\label{tab:F3-F4}
\end{minipage}
\end{table}

\begin{table}[htbp]

\begin{minipage}[t]{0.48\textwidth}
\centering
\begin{tabular}{@{}llll@{}}
\toprule
\textbf{ID} & \textbf{} & \textbf{HCTPS-GA} & \textbf{GA} \\ \midrule
F5 & \begin{tabular}[c]{@{}l@{}}Mean\\ Best\\ Worst\\ Median\\ St. Dev\end{tabular} & 
\begin{tabular}[c]{@{}l@{}}1.13712E-10\\ 9.5848E-11\\ 1.26901E-10\\ 1.14559E-10\\ 7.79472E-12\end{tabular} & 
\begin{tabular}[c]{@{}l@{}}1.1122E-10\\ 8.9499E-11\\ 1.2968E-10\\ 1.1166E-10\\ 1.1879E-11\end{tabular} \\
F6 & \begin{tabular}[c]{@{}l@{}}Mean\\ Best\\ Worst\\ Median\\ St. Dev\end{tabular} & 
\begin{tabular}[c]{@{}l@{}}2.840347319\\ 2.840347319\\ 2.840347319\\ 2.840347319\\ 4.07358E-12\end{tabular} & 
\begin{tabular}[c]{@{}l@{}}5.1220E+0\\ 2.8251E+0\\ 3.2955E+1\\ 2.8251E+0\\ 6.8450E+0\end{tabular} \\
\bottomrule
\end{tabular}
\caption{Performance Comparison (F5-F6)}
\label{tab:F6-F7}
\end{minipage}
\hfill
\begin{minipage}[t]{0.48\textwidth}
\centering
\begin{tabular}{@{}llll@{}}
\toprule
\textbf{ID} & \textbf{} & \textbf{HCTPS-GA} & \textbf{GA} \\ \midrule
F7 & \begin{tabular}[c]{@{}l@{}}Mean\\ Best\\ Worst\\ Median\\ St. Dev\end{tabular} & 
\begin{tabular}[c]{@{}l@{}}0.500000001\\ 0.500000001\\ 0.500000001\\ 0.500000001\\ 7.64014E-11\end{tabular} & 
\begin{tabular}[c]{@{}l@{}}6.9444E+4\\ 5.4066E+4\\ 8.4695E+4\\ 7.0732E+4\\ 1.0067E+4\end{tabular} \\
F8 & \begin{tabular}[c]{@{}l@{}}Mean\\ Best\\ Worst\\ Median\\ St. Dev\end{tabular} & 
\begin{tabular}[c]{@{}l@{}}0.459947694\\ 0.459947694\\ 0.459947694\\ 0.459947694\\ 2.64002E-11\end{tabular} & 
\begin{tabular}[c]{@{}l@{}}3.5113E+12\\ 6.6581E+3\\ 2.7423E+13\\ 4.6568E+10\\ 8.1503E+12\end{tabular} \\
\bottomrule
\end{tabular}
\caption{Performance Comparison (F7-F8)}
\label{tab:F7-F8}
\end{minipage}
\vskip\floatsep

\begin{minipage}[t]{0.48\textwidth}
\centering
\begin{tabular}{@{}llll@{}}
\toprule
\textbf{ID} & \textbf{} & \textbf{HCTPS-GA} & \textbf{GA} \\ \midrule
F9 & \begin{tabular}[c]{@{}l@{}}Mean\\ Best\\ Worst\\ Median\\ St. Dev\end{tabular} & 
\begin{tabular}[c]{@{}l@{}}4.43632E-11\\ 6.56031E-12\\ 9.47957E-11\\ 3.7284E-11\\ 3.07707E-11\end{tabular} & 
\begin{tabular}[c]{@{}l@{}}3.3533E-1\\ -4.1880E-1\\ 4.9625E-1\\ 4.6182E-1\\ 2.5453E-1\end{tabular} \\
F10 & \begin{tabular}[c]{@{}l@{}}Mean\\ Best\\ Worst\\ Median\\ St. Dev\end{tabular} & 
\begin{tabular}[c]{@{}l@{}}2\\ 2\\ 2\\ 2\\ 3.27683E-11\end{tabular} & 
\begin{tabular}[c]{@{}l@{}}1.2824E+8\\ 4.1754E+3\\ 6.4375E+8\\ 6.8138E+7\\ 1.6987E+8\end{tabular} \\
\bottomrule
\end{tabular}
\caption{Performance Comparison (F9-F10)}
\label{tab:F9-F10}
\end{minipage}
\hfill
\begin{minipage}[t]{0.48\textwidth}
\centering
\begin{tabular}{@{}llll@{}}
\toprule
\textbf{ID} & \textbf{} & \textbf{HCTPS-GA} & \textbf{GA} \\ \midrule
F11 & \begin{tabular}[c]{@{}l@{}}Mean\\ Best\\ Worst\\ Median\\ St. Dev\end{tabular} & 
\begin{tabular}[c]{@{}l@{}}0\\ 0\\ 0\\ 0\\ 0\end{tabular} & 
\begin{tabular}[c]{@{}l@{}}1.7259E+1\\ 1.0755E+1\\ 2.2015E+1\\ 1.7394E+1\\ 2.7755E+0\end{tabular} \\
F12 & \begin{tabular}[c]{@{}l@{}}Mean\\ Best\\ Worst\\ Median\\ St. Dev\end{tabular} & 
\begin{tabular}[c]{@{}l@{}}0\\ 0\\ 0\\ 0\\ 0\end{tabular} & 
\begin{tabular}[c]{@{}l@{}}7.0787E+4\\ 4.6881E+4\\ 9.2978E+4\\ 6.9242E+4\\ 1.0849E+4\end{tabular} \\
\bottomrule
\end{tabular}
\caption{Performance Comparison (F11-F12)}
\label{tab:F11-F12}
\end{minipage}

\vskip\floatsep

\begin{minipage}[t]{1\textwidth}
\centering
\begin{tabular}{@{}llll@{}}
\toprule
\textbf{ID} & \textbf{} & \textbf{HCTPS-GA} & \textbf{GA} \\ \midrule
F13 & \begin{tabular}[c]{@{}l@{}}Mean\\ Best\\ Worst\\ Median\\ St. Dev\end{tabular} & 
\begin{tabular}[c]{@{}l@{}}5.42E-20\\ 3.99E-20\\ 7.49E-20\\ 5.54E-20\\ 9.02E-21\end{tabular} & 
\begin{tabular}[c]{@{}l@{}}6.8974E+4\\ 5.8835E+4\\ 7.9979E+4\\ 6.8824E+4\\ 6.4668E+3\end{tabular} \\
F14 & \begin{tabular}[c]{@{}l@{}}Mean\\ Best\\ Worst\\ Median\\ St. Dev\end{tabular} & 
\begin{tabular}[c]{@{}l@{}}1.68E-10\\ 1.47E-10\\ 1.84E-10\\ 1.68E-10\\ 1.18E-11\end{tabular} & 
\begin{tabular}[c]{@{}l@{}}2.1332E+1\\ 2.1121E+1\\ 2.1544E+1\\ 2.1363E+1\\ 1.0935E-1\end{tabular} \\
\bottomrule
\end{tabular}
\caption{Performance Comparison (F13-F14)}
\label{tab:F13-F14}
\end{minipage}
\end{table}

\newpage
\subsection{Discussion}
Throughout these experiments, various configurations of the SSCP were employed during the local search phase, with each configuration corresponding to a distinct sequence-generative rule. For clarity, details of experiments involving sequence-generative rules that yielded unsatisfactory results in comparison to the global search phase are omitted. Instead, this section focuses on configurations that proved effective for each problem instance. The reason is, unsuccessful attempts in heuristics cannot provide us with information according to which one can infer any result general enough about the function in question or sufficient to theorization. The results of the global search phase are presented in table (\ref{tab:HCTPS-GA-RES}).

For the local search phase, the successful SSCP configuration involved the following steps:

\begin{enumerate}
\item \textbf{Lower-dimensional subdivision (dim=3)}: The cube $I=[-100, 100]^3$, was subdivided into $2^3 = 8$ equal-volume subcubes by splitting each dimension at its midpoint. The resulting subcubes were arranged into a sequence based on a clockwise traversal of the lower layer, followed by the upper layer of $I$ . Starting at one corner of the lower layer, the sequence progresses as:
$$
\{I_n\}_{n=1}^8 =
\begin{aligned}
&\big\{
[-100, 0] \times [-100, 0] \times [-100, 0], 
[-100, 0] \times [-100, 0] \times [0, 100], \\
&[-100, 0] \times [0, 100] \times [-100, 0], [-100, 0] \times [0, 100] \times [0, 100], \\
&[0, 100] \times [-100, 0] \times [-100, 0], [0, 100] \times [-100, 0] \times [0, 100], \\
&[0, 100] \times [0, 100] \times [-100, 0], [0, 100] \times [0, 100] \times [0, 100]
\big\}.
\end{aligned}
$$

Local search was performed within each subcube to identify the region with the best results. For example, for F1, the subcube:
   $$
   \mathcal{S}_3 = [0, 100] \times [-100, 0] \times [0, 100],
   $$ was selected
   
\item \textbf{Extension to higher dimensions}: The selected subcube was extended to 30 dimensions by cyclically replicating its bounds across all dimensions. For FI, for any dimension $x_i$, the bounds were:
   $$
   x_i \in
   \begin{cases}
   [0, 100], & \text{if } (i-1) \bmod 3 = 0, \\
   [-100, 0], & \text{if } (i-1) \bmod 3 = 1, \\
   [0, 100], & \text{if } (i-1) \bmod 3 = 2.
   \end{cases}
   $$
This resulted in the 30-dimensional search subcube:
   $$
   \mathcal{S}_{30} = \prod_{k=1}^{10} \mathcal{S}_3,
   $$
   where $\mathcal{S}_3$ was replicated 10 times to cover all dimensions.

\item \textbf{Scaling}: For most problems, the 30-dimensional subcube was scaled by a factor of $(1/2)^m$, where $m > 0$ is an integer. For F1, we used $m = 80$, resulting in the final search space:
   $$
   \mathcal{S}_{30,\text{scaled}} = \left( \frac{1}{2} \right)^{80} \times \mathcal{S}_{30}.
   $$

\end{enumerate}

The following tables(\ref{tab:selected_subcubes}, \ref{tab:scaling_factors}) present the selected subcube similar to $\mathcal{S}_3$ to F1, for each test function, and the scaling factors used if any respectively.

\begin{table}[htbp]
\centering
\small
\caption{Selected Subcubes for Each Test Function}
\label{tab:selected_subcubes}
\begin{tabular}{|c|c|}
\hline
\textbf{ID} & \textbf{Selected Subcube (\(\mathcal{S}_3\))} \\ \hline
F1 & \([0, 100] \times [-100, 0] \times [0, 100]\) \\ \hline
F2 & \([0, 100] \times [-100, 0] \times [-100, 0]\) \\ \hline
F3 & \([0, 100] \times [0, 100] \times [0, 1000]\) \\ \hline
F4 & \([-100, 0] \times [0, 100] \times [0, 100]\) \\ \hline
F5 & \([0, 100] \times [0, 100] \times [-100, 0]\) \\ \hline
F6 & \([0, 100] \times [-100, 0] \times [0, 100]\) \\ \hline
F7 & \([-100, 0] \times [-100, 0] \times [-100, 0]\) \\ \hline
F8 & \([0, 100] \times [0, 100] \times [-100, 0]\) \\ \hline
F9 & \([-100, 0] \times [0, 100] \times [-100, 0]\) \\ \hline
F10 & \([-100, 0] \times [0, 100] \times [0, 100]\) \\ \hline
F11 & \([-100, 0] \times [-100, 0] \times [-100, 0]\) \\ \hline
F12 & \([-100, 0] \times [0, 100] \times [-100, 0]\) \\ \hline
F13 & \([0, 100] \times [-100, 0] \times [0, 100]\) \\ \hline
F14 & \([0, 100] \times [-100, 0] \times [-100, 0]\) \\ \hline
\end{tabular}
\label{tab:subcubes}
\end{table}

\begin{table}[htbp]
\centering
\small
\caption{Scaling Factors Applied in the Local Search Phase}
\label{tab:scaling_factors}
\begin{tabular}{|c|c|}
\hline
\textbf{ID} & \textbf{Scaling Factor (\((1/2)^m\))} \\ \hline
F1  & \((1/2)^{80}\) \\ \hline
F2  & \((1/2)^{40}\) \\ \hline
F3  & None \\ \hline
F4  & \((1/2)^{10}\) \\ \hline
F5  & None \\ \hline
F6  & \((1/2)^{20}\) \\ \hline
F7  & \((1/2)^{40}\) \\ \hline
F8  & \((1/2)^{40}\) \\ \hline
F9  & \((1/2)^{40}\) \\ \hline
F10 & \((1/2)^{40}\) \\ \hline
F11 & \((1/2)^{40}\) \\ \hline
F12 & \((1/2)^{40}\) \\ \hline
F13 & \((1/2)^{40}\) \\ \hline
F14 & \((1/2)^{40}\) \\ \hline
\end{tabular}
\label{tab:scales}
\end{table}

\newpage
As seen in Tables (\ref{tab:HCTPS-GA-RES}, \ref{tab:F1-F2}, \ref{tab:F3-F4}, \ref{tab:F6-F7}, \ref{tab:F7-F8}, \ref{tab:F9-F10}, \ref{tab:F11-F12}, \ref{tab:F13-F14}), the results of the integration of the GA within the HCTPS framework outperforms the standalone GA in terms of best objective values and statistical measures. Through the HCTPS-GA, we managed to find solutions that are within a tight vicinity of the known global optimum for each objective function, and even achieved the exact global optimum for F11 and F12. However, it is important to emphasize that the \textit{success} of reaching a tight neighborhood of the global optimum or finding the global optimum itself for a specific problem within the HCTPS depends on finding the \textit{appropriate} sub-cube $J$ for the search algorithm $g$ in use, and since $J$ cannot be known a priori especially when dealing with black box optimization problems, the user may need to experiment with multiple sequence-generative rules before identifying an optimal SSCP configuration in favor of finding the right sub-cube.

\section{HCTPS Assessment}
\label{sec:assess}
We evaluate HCTPS in light of the criteria that define a systematic resolution to the exploration-exploitation dilemma in the introduction section (\ref{sec: intro}), namely:

\begin{enumerate}
\item Decoupled and independent control of exploration and exploitation
\item Providing explicit and adaptive regulation of exploration and exploitation
\item Theoretical and computational feasibility 
\end{enumerate}

In addition, we also consider the cost–benefit implications of adopting HCTPS in practice. Below, we summarize how HCTPS addresses each point and examine its comparative advantages over the current approach.

\subsection{Decoupled and Independent Control}
A principal contribution of HCTPS is that it decouples exploration from exploitation by shifting the regulatory burden to an external parameter—the SSCP. Concretely:

\begin{itemize}
\item \textbf{Global Phase}: The EA’s inherent exploitation mechanisms (e.g., its population convergence tendencies, mutation/selection synergy, etc.) remain intact and need not be modified or “loosened” to increase exploration. Instead, the algorithm simply executes as usual across the entire search cube 

\item \textbf{Local Phase}: Exploration is selectively and additively intensified in chosen sub-cubes $\{I_i\}_{i=1}^n$ of $I$. By letting the user dynamically specify or adjust which sub-cubes to explore and to what extent, HCTPS ensures that raising exploration does not implicitly undermine the EA’s convergence properties.
\end{itemize}

This external and human-centered mechanism fundamentally differs from the traditional approach (i.e., re-tuning internal operator parameters). Exploration and exploitation thus become governed by distinct, independently adjustable mechanisms: the EA’s built-in exploitation potential, on the one hand, and the SSCP-based search-space relocation, on the other.

\subsection{Explicit and Adaptive Regulation}
An equally crucial requirement for a systematic resolution is that exploration be regulated explicitly and be adaptive to real-time observations. HCTPS satisfies these in the following ways:

\begin{enumerate}
\item \textbf{Explicit parameter (SSCP)}: The SSCP is the dominant and only parameter the user needs to focus on to modulate exploration. Rather than distributing tuning efforts among mutation rates, crossover probabilities, or other interdependent parameters, the user exerts direct control over sub-cube selection and size (including possible scaling factors).
\item \textbf{Adaptive local search}: The user can progressively generate additional sub-cubes $\{I_i\}$ or refine existing ones based on observed partial results (e.g., fitness improvement rates, solution clustering, .. etc). This selective local-phase configuration (Step 2 and Step 3 in procedure (\ref{algo:HCTPS}) thus adapts the exploration intensity to real-time insights.
\end{enumerate}

By centralizing exploration decisions into one external “knob,” HCTPS avoids entangled parameter interdependencies, thereby enabling a principled approach to increasing or reducing exploration as the search unfolds.

\subsection{Theoretical and computational feasibility}
HCTPS meets the third requirement for a systematic resolution by remaining algorithm-agnostic, not imposing exponential expansions in representation size, nor mandating extensive operator re-tuning:

\begin{itemize}
\item \textbf{Algorithm-agnostic}: HCTPS is a procedural add-on that can be combined with any evolutionary or non-evolutionary search algorithm $g$. The only assumption is that $g$ can run on a given subregion of the search space.
\item \textbf{No exponential increase in representation set}: Unlike attempts to boost exploration by lengthening encodings or using prohibitively large populations, HCTPS redefines which portion of $I$, the algorithm sees at each local phase, avoiding unbounded expansions in representation cardinality.
\item \textbf{No exhaustive parameter sweeps}: By focusing all exploration tuning on the SSCP, HCTPS circumvents the need to balance multiple, unintelligibly interdependent parameters. This single-parameter approach is less prone to combinatorial parameter inflation.
\end{itemize}

Moreover, the theoretical guarantee in Theorem \ref{thm: hctps} establishes that HCTPS strictly improves exploration coverage over the traditional approach, provided the chosen sub-cubes in the local phase are sufficiently diverse.

\subsection{Cost–Benefit Analysis}
Although the user (i.e., decision-maker) invests time in constructing or refining an SSCP-based sequence of sub-cubes, this investment replaces the extensive trial-and-error that typically arises when tuning many internal EA parameters simultaneously:

\begin{enumerate}
\item \textbf{Focused tuning effort}: All exploration-specific customizations are consolidated into one parameter class (SSCP). This spares the user from contending with unintelligible interdependencies among operators, probabilities, and population dynamics.

\item \textbf{Incremental refinement}: Even when certain SSCP configurations fail, they often yield search-space insights regarding which subregions are not promising. This is arguably more interpretable than concluding that “some parameter triple 
$(N, p_c, p_m)$ was suboptimal” in a black-box sense.

\item \textbf{Parallelizability}: Since each sub-cube can be handled in an isolated local phase, the user may run local-phase searches in parallel if computational resources permit, further reducing overhead.
\end{enumerate}

Ultimately, the gains in exploration efficacy and interpretability of the local-phase results typically outweigh the cost of specifying or updating SSCP rules.

\section{Summary, and Key Takeaways}
\label{sec:conclusions}

\subsection{Summary} 
The fundamental obstacle identified in the current approach practiced in the literature in dealing with the exploration-exploitation dilemma arises from two interacting issues:
\begin{enumerate}
\item The fixed representation set $X$ size in which the EA operates
\item The unintelligible interdependencies between its traditional parameters
\end{enumerate}

(1) alone implies that, at best, the algorithm will only be able to unlock a subset from $I$ with the same cardinality of $X$, and whose cardinality is a function of the pre-determined parameter $L$ (solutions bit-encoding length). While (2) alone implies that the collective contribution of the internal mechanics of the EA is not predictable. (1) and (2) together makes it very reasonable to overlook potential promising regions or solutions if they happen to not exit in the subset we are to unlock. Increasing exploration under (1) and (2) in the current approach necessitates increasing the size of $X$ and population size $N$ as perminilaries and the consequences is a huge computational burden.

To avoid such consequences, we proposed a two-phase search framework to which the human-user serves as a meta-paramter, thus the name "human-centered two-phase search" (HCTPS). This framework allows for the free adaptive adjustment of exploration without affecting the inherent exploitation of the algorithm in use, thus make it possible to reach the proper exploration to exploitation ratio that fit the specific problem under consideration and the underlying algorithm in use, and while being theoretically and computationally feasible, and thus resolves the exploration exploitation dilemma.

In section (\ref{sec: HCTPS}), we detailed the search process within the proposed framework, which restructures the search into two consecutive phases: a global search phase followed by a local search phase. In the first, the EA operates over the entire search cube $I$ for $k$ iterations, and in this the EA distributes its inherent exploration-exploitation potential over $I$. In the later, the search is carried out over \textit{selected} partitions "sub-cubes" of $I$ each for the same $k$ iterations, and in this the EA distributes its inherent exploitation-explortion potential over different-in-size sub-regions of $I$, and the latter phase continues until satisfactory results are identified. The rigorous implementation of the local search phase is facilitated through the introduction of SSCP which follows a human-centered tuning strategy (see section \ref{subsec:par}), and a compact description of the HCTPS was provided by procedure (\ref{algo:HCTPS}).

\subsubsection{HCTPS Effect on EAs as Global Optimizers}
Resolving the exploration-exploitation dilemma in EAs is a prerequisite for enhancing their utility as global optimization tools—the latter cannot follow without the first. Therefore, applying an arbitrary EA (i.e., any EA with some fixed parameter values) within the HCTPS and comparing against the same EA as standalone serves as a nice way to illustrate its concrete implementation, and demonstrate its effectiveness against the current approach (i.e, focusing all tunning efforts on the SSCP vs distribution over traditional EA parameters), and this was carried out in section (\ref{sec:experiments}) with the canonical GA as an algorithm choice.

The aim of the experiment is not to serve as a tie-breaker to the framework effectiveness which means an unsuccessful SSCP configuration doesn't imply that the framework is defective but rather that the specific SSCP configuration is not the right one for the specific algorithm in use and the problem it is applied to. The framework’s guaranteed resolution of the dilemma in question, and thereby increasing the chances of hitting a tight approximation of global optima is supported by its theoretical and conceptual foundations, as presented in section (\ref{subsec: hctps}).

\subsection{Key Takeaways}
The key takeaways from this work can be summarized as follows:  

\begin{enumerate}

\item \textbf{Reconciling exploration–exploitation}: Within the HCTPS framework, exploration and exploitation are not inherently conflicting because they are decoupled. This allows for the maximization of the first without diminishing the latter, a necessary condition for effective search performance.

\item \textbf{Significantly enhanced coverage}: Embedding any EA within HCTPS significantly increases search coverage and outperforms the same EA without HCTPS. The theoretical foundation shows that decoupling exploration in this way \emph{strictly expands} the set of points the algorithm can effectively sample, thus reducing the chances of missing promising niches.

\item \textbf{Single-parameter focus}: A key advantage of HCTPS is that all exploration-related tuning efforts are directed towards a \emph{single} parameter (the SSCP), thereby removing many of the unintelligible, compounded interactions among traditional EA parameters.

\item \textbf{Adaptability and trial-and-error}: While the fine-tuning of the SSCP may involve an element of trial and error, this is neither unique to HCTPS nor a hindrance to its adoption. EAs universally require trial and error experimentation to tailor parameter values to specific problem contexts. The human-centered SSCP, however, offers much more valuable returns regarding maximizing exploration without introducing unnecessary randomization to the structure of the algorithm in use in addition to its adaptability, simple tuning process and intuitive implementation. Moreover, the ability to parallelize sub-cube testing provides a practical avenue for alleviating the computational burden, enabling concurrent exploration of multiple regions and expediting the search process.

\item \textbf{Broad applicability}:  The HCTPS framework applicability is not limited to EAs; the general idea of “decouple and externally direct exploration” applies to any search method, not just EAs. Whenever an algorithm’s default parameters alone fail to explore sufficiently, an external “search-space control” layer can fill that gap.

\item In closing, the HCTPS provides an overarching reconcilatory framework that unifies exploration and exploitation without compromising either one. By disentangling these two processes—and placing exploration under dynamic human-centered control—the framework has the potential to offer a robust, extensible means of improving global search outcomes across a wide spectrum of search computational problems prone to the exploration exploitation dilemma.
\end{enumerate}

\section{Future Work}
Future research could potentially focus on developing interactive systems where humans and automated agents work in collaboration to refine sequence-generative rules for search space partitioning. One promising direction involves designing user interfaces that provide real-time visualization of the search space and the current partitioning strategy. Such interfaces would display performance metrics and spatial distributions of explored subcubes, enabling users to quickly identify regions of interest or redundancy.

\section*{Acknowledgments}  
I sincerely thank the organizers SIAM-MDS24 for the opportunity to present and discuss early phases of this work, professor Rana Dajani (Hashemite University) for clarifying evolutionary biology concepts which helped me refine some paragraphs in the introduction section, and professor Thomas Bäck (Leiden University) for his comments on the early version of this manuscript, which inspired improving clarity in the presentation of its ideas in this version.

\nocite{*}
\bibliographystyle{unsrt}  
\bibliography{RTMLR.bib}

\end{document}